\documentclass[11pt, a4paper, copyright]{google}

\usepackage{tikz}
\usepackage{multicol}
\usetikzlibrary{positioning,arrows.meta,shapes.geometric,fit,backgrounds,calc,decorations.pathreplacing}

\usepackage[authoryear, sort&compress, round]{natbib}
\makeatletter
\renewcommand{\absfont}{\normalfont\linespread{1.2}\fontsize{11}{12}\selectfont}

\definecolor{zjublue}{RGB}{0,63,136}      
\definecolor{zjulightblue}{RGB}{0,159,227} 

\usepackage[colorlinks = true,
            linkcolor = zjublue,
            urlcolor  = zjulightblue,
            citecolor = zjublue,
            anchorcolor = zjublue]{hyperref}
\usepackage[misc]{ifsym}
\usepackage{minitoc}
\usepackage{cleveref}
\usepackage{hyperref}
\usepackage{subcaption}
\usepackage{booktabs}
\usepackage{fontawesome5}
\usepackage{amsmath}
\usepackage{mathtools}
\usepackage{amssymb}
\usepackage{graphicx}
\usepackage{algorithmic}
\usepackage{algorithm}
\usepackage{tcolorbox}
\usepackage{verbatim}
\usepackage{makecell}
\usepackage{array}
\usepackage{multirow}
\usepackage{xurl}
\usepackage{amsthm}
\usepackage{adjustbox}
\usepackage{caption}
\usepackage{xcolor}
\usepackage{wrapfig}
\usepackage{pifont}
\usepackage{enumitem}

\usepackage{colortbl}
\usepackage{xcolor}

\usepackage{titlesec}
\titlespacing*{\paragraph}{0pt}{0pt}{1em}

\definecolor{headerblue}{RGB}{180, 210, 240}
\definecolor{rowgray}{RGB}{248, 248, 248}
\definecolor{closedrow}{RGB}{255, 245, 220}
\definecolor{groupblue}{RGB}{220, 235, 250}
\definecolor{successgreen}{RGB}{46, 139, 87}
\definecolor{failred}{RGB}{200, 60, 60}
\definecolor{groupheader}{RGB}{230, 230, 230}

\usepackage[titles]{tocloft}
\definecolor{tocsubsec}{HTML}{444444}
\definecolor{tocsubsubsec}{HTML}{666666}

\usepackage{soul}

\usepackage{listings}
\tcbuselibrary{listings,skins,breakable}

\lstdefinestyle{prompt}{
  basicstyle=\ttfamily\footnotesize,
  breaklines=true,
  breakatwhitespace=true,
  columns=fullflexible,
  keepspaces=true,
  showstringspaces=false,
  postbreak=\mbox{\textcolor{gray}{$\hookrightarrow$}\space}
}

\definecolor{accent1}{HTML}{2E6DA4}
\definecolor{accent2}{HTML}{D9534F}
\definecolor{accent3}{HTML}{5CB85C}
\definecolor{lightblue}{RGB}{173,216,230}
\definecolor{lightorange}{RGB}{255,213,170}
\definecolor{lightgreen}{RGB}{176,226,176}
\definecolor{lightyellow}{RGB}{255,255,204}
\definecolor{lightgray}{RGB}{220,220,220}
\definecolor{lightpurple}{RGB}{221,160,221}
\definecolor{lightred}{RGB}{255,182,193}
\definecolor{gray60}{gray}{0.6}
\definecolor{accent4}{HTML}{9B59B6}

\uselogo{}

\makeatletter
\renewcommand{\maketitle}{\bgroup\setlength{\parindent}{0pt}
  \begin{adjustwidth}{0pt}{24pt}
    \begin{center}
      {\titlefont \@title\par}%
      \vskip11pt
      {\@author\par}%
      \vskip20pt%
    \end{center}
  \end{adjustwidth}
  \egroup
  {\abscontent}%
  \thispagestyle{firststyle}
}
\makeatother

\AtBeginDocument{%
  \fancypagestyle{firststyle}{%
    \fancyhead[L]{\includegraphics[height=27pt]{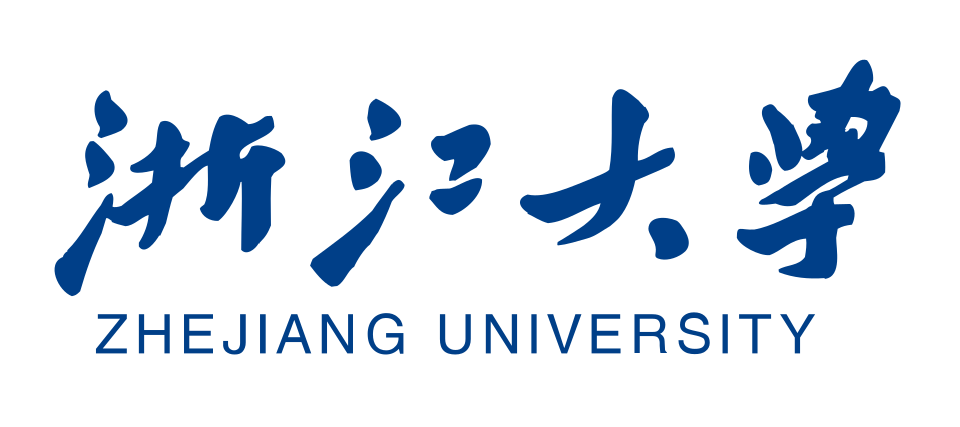}}%
    \fancyhead[R]{\includegraphics[height=27pt]{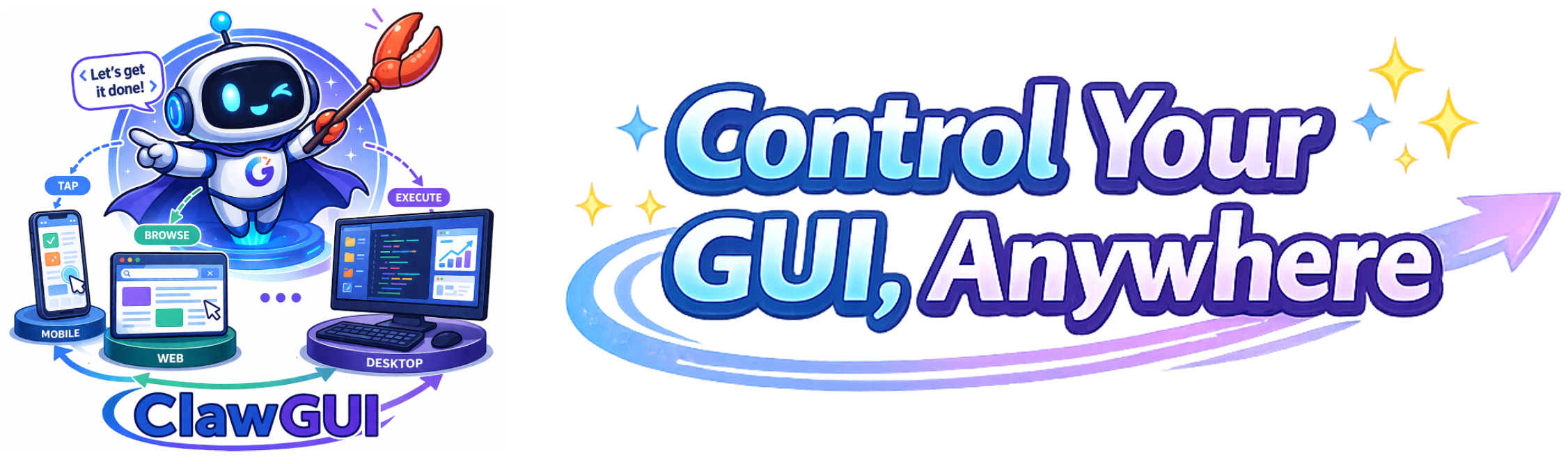}}%
    \fancyhead[C]{}%
    \fancyfoot[L]{%
  \footerfont
  $^{*}$Core contribution;\quad $^{\dagger}$Corresponding authors;\quad
  \ifdefined\correspondingauthor
    \if\relax\the\correspondingauthor\relax
    \else
      {\itshape \the\correspondingauthor}%
    \fi
  \fi
  \\
  \ifthenelse{\boolean{copyright}}{\copyrightext}{}%
}%
    \fancyfoot[R]{\footerfont\bfseries\relax}%
    \fancyfoot[C]{\footerfont\bfseries\relax}%
  }%
}

\title{ClawGUI: A Unified Framework for Training, Evaluating, and Deploying GUI Agents}

\correspondingauthor{Main contact: syl@zju.edu.com}
\author{%
  \textbf{Fei Tang}$^{*}$,~~
  \textbf{Zhiqiong Lu}$^{*}$,~~
  \textbf{Boxuan Zhang},~~
  \textbf{Weiming Lu},\\
  \textbf{Jun Xiao},~~ 
  \textbf{Yueting Zhuang},~~ 
  \textbf{Yongliang Shen}$^{\dagger}$~~
  \vspace{2pt}\\
  Zhejiang University~~ 
  \\
  \texttt{\{flysugar, syl\}@zju.edu.cn} \\
  \vspace{2pt}
  \faGithub~\href{https://github.com/zju-real/ClawGUI}{\texttt{https://github.com/zju-real/ClawGUI}}
  \vspace{2pt}
  \\
  ~~
  \faGlobe~\href{https://zju-real.github.io/ClawGUI-Page/}{\texttt{https://zju-real.github.io/ClawGUI-Page}}
}

\begin{abstract}
\vspace{-4mm}
\textbf{Abstract.} 
GUI agents drive applications through their visual interfaces instead of programmatic APIs, interacting with arbitrary software via taps, swipes, and keystrokes, reaching a long tail of applications that CLI-based agents cannot.
Yet progress in this area is bottlenecked less by modeling capacity than by the absence of a coherent full-stack infrastructure: 
online RL training suffers from environment instability and closed pipelines, evaluation protocols drift silently across works, and trained agents rarely reach real users on real devices. 
We present \textbf{ClawGUI}, an open-source framework addressing these three gaps within a single harness. \textbf{ClawGUI-RL} provides the first open-source GUI agent RL infrastructure with validated support for both parallel virtual environments and real physical devices, integrating GiGPO with a Process Reward Model for dense step-level supervision. \textbf{ClawGUI-Eval} enforces a fully standardized evaluation pipeline across 6 benchmarks and 11+ models, achieving 95.8\% reproduction against official baselines. \textbf{ClawGUI-Agent} brings trained agents to Android, HarmonyOS, and iOS through 12+ chat platforms with hybrid CLI-GUI control and persistent personalized memory.
Trained end to end within this pipeline, \textbf{ClawGUI-2B} achieves 17.1\% Success Rate on MobileWorld GUI-Only, outperforming the same-scale MAI-UI-2B baseline by 6.0\%.
\end{abstract}

\begin{document}

\maketitle

\begin{figure}[h]
  \centering
  \includegraphics[width=1.0\linewidth]{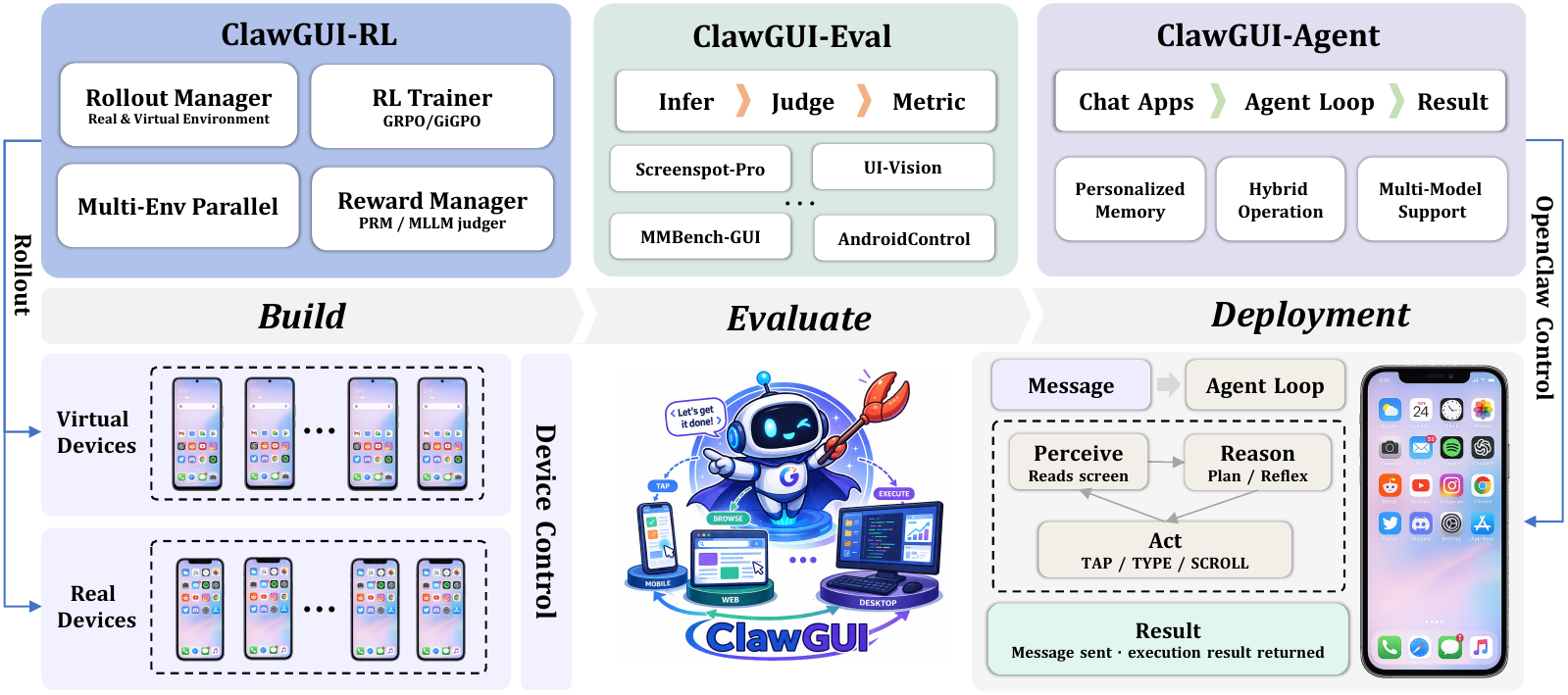}
\caption{Overview of ClawGUI, a unified open-source framework for GUI agent research and deployment. It integrates scalable online RL training (ClawGUI-RL) with outcome-based rewards and no human annotation, reproducible three-stage evaluation across 6 benchmarks and 11+ models (ClawGUI-Eval), and real-device deployment across Android, HarmonyOS, and iOS through 12+ chat platforms (ClawGUI-Agent).}
  \label{fig:framework}
\end{figure}

\newpage
\vspace{0.5em}
{
  \hypersetup{linkcolor=black}
  \setlength{\parskip}{0pt}
  \renewcommand{\contentsname}{\normalfont\large\bfseries Contents}
  \setcounter{tocdepth}{3}
  \begingroup
    \small
    \tableofcontents
  \endgroup
}

\newpage

\section{Introduction}
\label{sec:intro}

Graphical User Interfaces (GUIs) are the universal substrate through which humans interact with modern computing devices~\citep{tang2025surveymllmbasedguiagents,fu2023ufo2unifiedpretrainingframework,shen2023hugginggptsolvingaitasks,nakano2022webgptbrowserassistedquestionansweringhuman,hong2024cogagentvisuallanguagemodel,tang2025thinktwiceclickonce}. An agent that can perceive screen state and execute low-level interface actions such as tapping, swiping, and typing is, in principle, capable of operating any application on any device without requiring dedicated APIs or backend access~\citep{hu2024dawnguiagentpreliminary,lai2024autowebglmlargelanguagemodelbased,zhang2025apiagentsvsgui}. This generality has made GUI agents one of the most actively pursued directions toward end-to-end digital automation, with rapid progress over the past two years across grounding, navigation, and online reinforcement learning (RL)~\citep{mobileguirl,qin2025uitarspioneeringautomatedgui}.

\noindent Building a capable GUI agent, however, is not a single modeling problem but a full-stack engineering problem~\citep{luo2025guir1generalistr1style,lu2025uir1enhancingefficientaction,tang2025gui,mobileagent3}. A useful agent must be trained against realistic environments, evaluated under comparable conditions, and ultimately deployed to real devices where real users can benefit from it~\citep{mobileguirl,uitars2,liu2024autoglmautonomousfoundationagents}. Existing research has made meaningful progress on each of these fronts in isolation: GUI grounding models have steadily improved element localization accuracy~\citep{luo2025guir1generalistr1style,lu2025uir1enhancingefficientaction,liu2025infiguir1advancingmultimodalgui,tang2025gui}, navigation agents have extended task horizons~\citep{qin2025uitarspioneeringautomatedgui,uivenus}, and online reinforcement learning has begun to push policy quality beyond what static supervision alone can achieve~\citep{uivenus15}. Yet when one attempts to assemble these pieces into a working pipeline, the cracks between them become apparent. The community still lacks a unified framework in which training, evaluation, and deployment operate as a coherent whole, and this absence, rather than any single missing technique, is what currently bottlenecks practical progress.

\noindent We identify three concrete gaps that together define this bottleneck.

\paragraph{The training ecosystem for GUI agents remains largely closed.} Several recent systems report strong results from online RL training in virtual environments~\citep{uitars2,uivenus15,mobileagent35}, but none release the underlying infrastructure, leaving outside researchers unable to reproduce the setup or build upon it. Even where code exists, it is tied exclusively to emulator-based sandboxes, and training directly on physical devices, which is ultimately where agents must perform, remains essentially unexplored in the open literature~\citep{maiui}. The central engineering difficulty here is not the RL algorithm itself but environment management: emulators drift out of healthy states during long runs, real devices cannot expose system-level verification signals, and reward signals in long-horizon GUI tasks are sparse almost by construction.

\paragraph{Evaluation across GUI agent papers is badly misaligned.} GUI benchmarks appear straightforward on the surface, yet reported numbers across papers are rarely directly comparable~\citep{seed18,Gemini}. Prompt formatting, coordinate normalization conventions, image resolution, and sampling configuration each shift reported accuracy by several points, and these choices are often undocumented. The result is that the community has no shared baseline against which to measure true progress. A 2\% improvement on ScreenSpot-Pro~\citep{li2024screenspot-pro} may reflect a genuine advance, a favorable prompt, or simply a different resolution, and there is currently no way for a reader to tell.

\paragraph{The deployment loop from research to real users is broken.} Agents trained in research pipelines almost never reach end users. A recent line of work has explored CLI-based agent harnesses~\citep{openclaw,claudecode,clianything,opencli}, which offer precise control through structured commands but cover only a narrow slice of real applications. Meanwhile, systems that connect a trained GUI policy to real hardware, expose it through interfaces users already use in daily life, and maintain persistent personalization over time remain largely absent from the open ecosystem~\citep{zhang2023appagentmultimodalagentssmartphone,wang2024mobileagentv2mobiledeviceoperation,agashe2024agentsopenagentic}. Without this final link, the real-world value of GUI agents goes largely unverified.

\noindent Building on these insights, we present ClawGUI, an open-source framework designed to close all three gaps within a single coherent system. ClawGUI consists of three tightly integrated modules. ClawGUI-RL provides scalable online RL infrastructure with validated support for both Docker-based parallel Android emulators and real physical devices, integrating GiGPO~\citep{gigpo} together with a Process Reward Model that supplies dense step-level supervision to counteract the sparsity of outcome rewards in long-horizon GUI tasks. ClawGUI-Eval enforces a strict three-stage pipeline across 6 benchmarks and 11+ models, pinning every evaluation choice per model so that results become reproducible rather than nominally comparable. 
\textbf{ClawGUI-Agent} closes the loop from research to deployment, bringing trained agents to Android, HarmonyOS, and iOS through 12+ chat platforms, with a hybrid CLI-GUI control strategy that combines the precision of CLI with the universal coverage of GUI, and a persistent personalized memory system that allows the agent to adapt to individual users over time.

\noindent To validate the framework end to end, we train ClawGUI-2B entirely within the ClawGUI-RL pipeline. On MobileWorld GUI-Only, ClawGUI-2B achieves a Success Rate (SR) of 17.1\%, compared to 11.1\% for the same-scale MAI-UI-2B baseline, and also surpasses substantially larger untrained models such as Qwen3-VL-32B (11.9\%) and UI-Venus-72B (16.4\%). Within the same pipeline, replacing episode-level GRPO with step-level GiGPO yields a 2.6\% improvement (14.5\% → 17.1\%), directly confirming the value of dense credit assignment in GUI RL. On the evaluation side, ClawGUI-Eval achieves a 95.8\% reproduction rate against published baselines across 6 benchmarks and 11+ models.


\noindent Our main contributions are as follows:
\begin{itemize}
  \item We release \textbf{ClawGUI}, a unified open-source framework that integrates online RL training, standardized evaluation, and real-device deployment into a single pipeline for GUI agents.
  \item We release \textbf{ClawGUI-RL}, the first open-source GUI agent RL infrastructure with validated support for both large-scale parallel virtual environments and real physical devices, integrating GiGPO with a Process Reward Model for dense step-level supervision.
  \item We release \textbf{ClawGUI-Eval} together with all inference code and pre-computed predictions across 6 benchmarks and 11+ models, achieving a 95.8\% reproduction rate against official baselines and enabling reliable cross-paper comparison.
  \item We release \textbf{ClawGUI-Agent}, a production-ready deployment system that connects trained agents to real Android, HarmonyOS, and iOS devices through 12+ chat platforms with persistent personalized memory.
  \item We release \textbf{ClawGUI-2B}, trained end to end within ClawGUI-RL, which reaches 17.1\% on MobileWorld GUI-Only and outperforms the same-scale MAI-UI-2B baseline by 6.0 absolute points, validating the framework end to end.
\end{itemize}

\section{Related Work}

\subsection{GUI Agent Models: From Grounding to Navigation}
Early GUI agents relied on cascaded pipelines that combined off-the-shelf perception modules such as OCR~\citep{wang2024mobileagentautonomousmultimodalmobile,du2025unirec01bunifiedtextformula,feng_etal_2025_dolphin,feng2026dolphinv2universaldocumentparsing}, SAM~\citep{kirillov2023segment,ravi2024sam2segmentimages}, and set-of-marks prompting~\citep{yang2023setofmarkpromptingunleashesextraordinary,lu2024omniparserpurevisionbased} with a closed-source planner, a modular but error-accumulating design that precluded end-to-end optimization. As vision-language foundation models matured, end-to-end grounding became the dominant paradigm: SeeClick~\citep{cheng2024seeclickharnessingguigrounding}, UI-TARS~\citep{qin2025uitarspioneeringautomatedgui}, Aguvis~\citep{xu2024aguvis}, and UGround~\citep{gou2024navigatingdigitalworldhumans} showed that localization accuracy scales with data and model capacity, and subsequent work sharpened grounding further via RL-based coordinate rewards~\citep{lu2025uir1enhancingefficientaction,luo2025guir1generalistr1style,tang2025gui}. On top of stronger grounding, a second wave targets long-horizon navigation, splitting into modular pipelines that pair a grounding model with a proprietary planner~\citep{gou2024navigatingdigitalworldhumans,xu2024aguvis} and unified end-to-end policies that internalize perception and decision-making jointly~\citep{qin2025uitarspioneeringautomatedgui,maiui,uivenus15}. ClawGUI is orthogonal to this modeling axis: rather than proposing a new grounding or navigation model, it provides a shared harness in which both paradigms can be trained, evaluated, and deployed under consistent conditions.

\subsection{Online Reinforcement Learning for GUI Agents}
Collecting large-scale trajectory data for long-horizon GUI tasks is expensive, as each demonstration requires step-by-step execution, precise action annotation, and faithful environment replay~\citep{lin2025guirewalkmassivedatageneration,wang2026openclawrltrainagentsimply,mobileworld}. Online reinforcement learning offers an attractive alternative, letting the agent generate its own experience through direct environment interaction and optimize toward task success via outcome rewards. A rapidly growing line of work including MobileGUI-RL~\citep{mobileguirl}, ComputerRL~\citep{lai2025computerrlscalingendtoendonline}, MAI-UI~\citep{maiui}, UI-Venus-1.5~\citep{uivenus15}, and UI-TARS-2~\citep{uitars2} has shown that sandbox-based online training yields consistent gains beyond SFT. Yet three difficulties persist: reward signals are sparse over long action sequences, multi-step credit assignment is nontrivial, and infrastructure cost is substantial, demanding parallel simulation and robust episode management across heterogeneous applications. More importantly for the community, none of these works open-source their training infrastructure, and all are validated solely in virtual sandboxes, leaving real-device training almost entirely unexplored. ClawGUI-RL directly targets this gap by releasing an open-source infrastructure that handles environment management, dense step-level reward supervision, and validated training on both parallel emulators and real physical devices.

\subsection{Benchmarking and Reproducibility of GUI Agents}
A rich ecosystem of GUI benchmarks has emerged to measure grounding and navigation, including ScreenSpot-Pro~\citep{li2024screenspot-pro}, ScreenSpot-V2, UI-Vision, MMBench-GUI, OSWorld-G, and AndroidControl, alongside interactive suites such as MobileWorld~\citep{mobileworld}. These have become the de facto yardsticks for reporting GUI agent progress. In practice, however, reported numbers across papers are rarely directly comparable: prompt formatting, coordinate normalization, image resolution, sampling temperature, and post-processing rules interact in ways that shift reported accuracy by several points, and many of these choices are undocumented~\citep{seed18,Gemini,tang2025thinktwiceclickonce}. As a result, the community has no reliable shared baseline, and small reported improvements are often indistinguishable from configuration drift. Prior standardization efforts have focused on a single benchmark, been bundled with a particular training recipe, or released evaluation scripts without the inference predictions, making independent re-judging infeasible. ClawGUI-Eval differs in both scope and philosophy: it decouples evaluation into standardized inference, judging, and metric computation, pins all configuration choices per model, and releases inference outputs across 6 benchmarks and 11+ models, enabling the community to reproduce, re-judge, and extend published results without re-running expensive inference.

\subsection{Deploying GUI Agents to Real Users}
A capable GUI agent delivers value only when it reaches real users operating real devices. Driven by the success of OpenClaw~\citep{openclaw} and Hermes-Agent~\citep{hermes_agent}, a growing body of work has turned toward CLI-based agent harnesses~\citep{opencli,clianything,claudecode,anthropic_harness_design_2026,hermes_agent,minimax_m27_news_2026} as a deployment substrate. CLI execution is efficient and precise, yet carries fundamental limitations: many applications expose no programmatic interface~\citep{zhang2025apiagentsvsgui}, CLI operations are opaque to users who cannot observe or intervene in agent behavior~\citep{masbench}, and bypassing the visual layer forfeits the spatial grounding that makes agent actions interpretable~\citep{tang2025surveymllmbasedguiagents,fu2025manotechnicalreport}. GUI-based interaction addresses these limitations by operating directly on the screen and covering any application regardless of its underlying architecture, but introduces its own cost, since tasks resolved by CLI in a single call may require several sequential GUI actions~\citep{jiang2025appagentxevolvingguiagents,Magentic-UI}. Existing research deployments also tend to stop at demo notebooks or isolated Android controllers, leaving cross-platform coverage and persistent personalization largely unaddressed. ClawGUI-Agent closes this gap through a hybrid harness that leverages CLI efficiency where interfaces permit and falls back to GUI control where they do not, connects trained agents to Android, HarmonyOS, and iOS through 12+ chat platforms, and integrates a persistent personalized memory system that enables the agent to adapt to individual users over time.

\section{ClawGUI}

\subsection{System Overview}
We introduce \textbf{ClawGUI}, a unified framework designed to cover the complete lifecycle of GUI agent development. As illustrated in Figure~\ref{fig:framework}, ClawGUI consists of three tightly integrated modules: \textbf{ClawGUI-RL} for scalable online RL training, \textbf{ClawGUI-Eval} for standardized and reproducible evaluation, and \textbf{ClawGUI-Agent} for real-device deployment and human interaction.


\begin{figure}[t]
  \centering
  \includegraphics[width=0.99\linewidth]{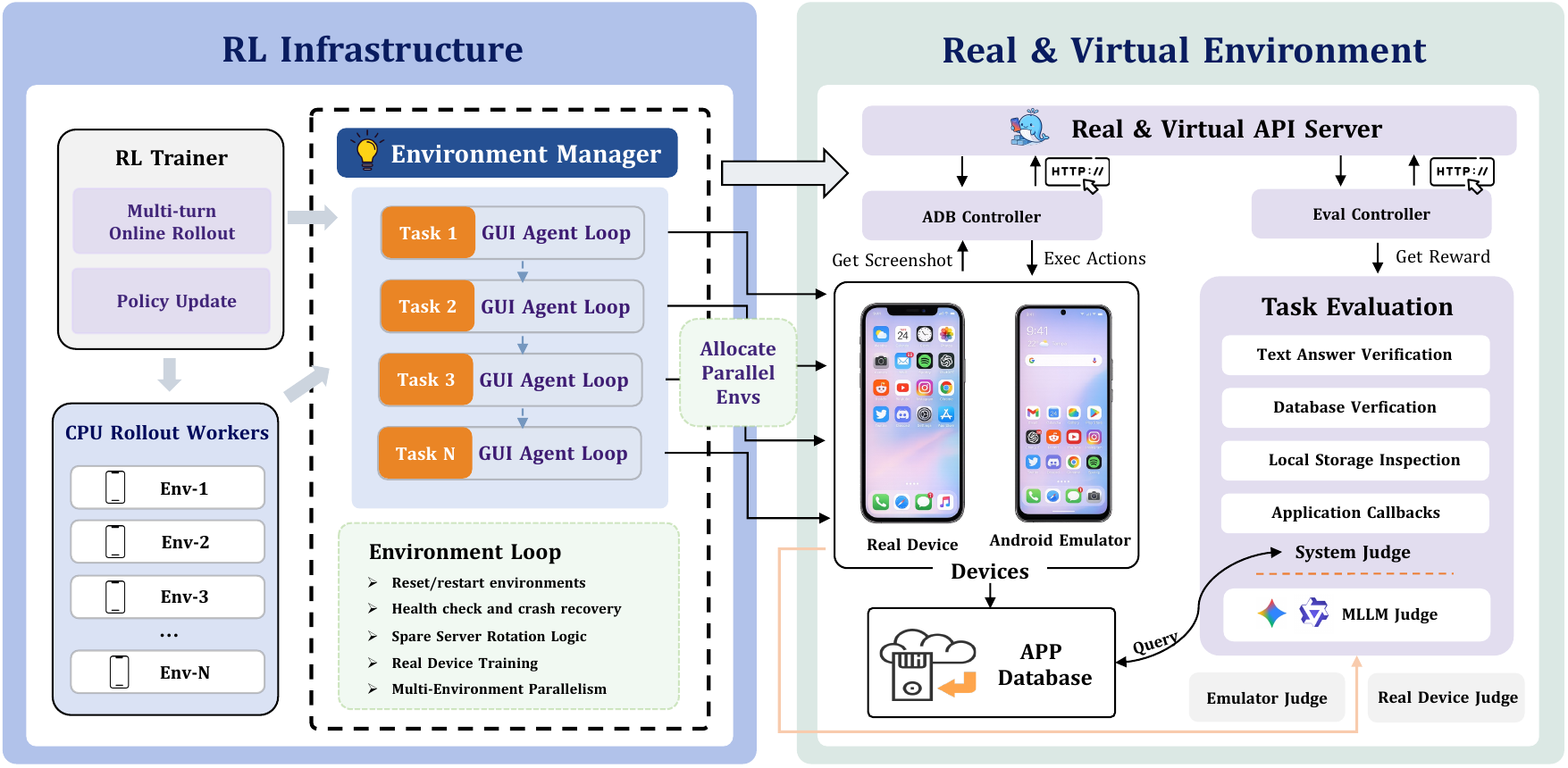}
  \caption{Overview of ClawGUI-RL, consisting of an RL Infrastructure 
and a Real \& Virtual Environment backend. The Environment Manager 
orchestrates multi-task parallel rollouts across real devices and 
Android emulators, with built-in health checking, crash recovery, 
and spare server rotation. Task evaluation combines system-level 
verification with MLLM-as-judge to provide robust reward signals 
for both virtual and real device training.}
  \label{fig:clawgui-rl}
\end{figure}

\subsection{ClawGUI-RL: Scalable Online RL Training}
GUI tasks are inherently sequential decision-making problems that require agents to learn through real environment interaction rather than static supervision alone. Despite growing interest in online RL for GUI agents, the community lacks an open-source infrastructure that is both scalable and validated on real physical devices. To this end, we build ClawGUI-RL to provide end-to-end support from environment management and reward design to policy optimization.

\subsubsection{Environment Manager}

Stable and scalable environment management is a prerequisite for online RL training on GUI tasks. As shown in Figure~\ref{fig:clawgui-rl}, ClawGUI-RL abstracts all device backends behind a unified interface, allowing virtual environments and physical devices to be used interchangeably within the same training loop.

\noindent\textbf{Virtual Environment.} ClawGUI-RL launches dozens of Docker-based Android emulators in parallel via MobileWorld~\citep{mobileworld}, each exposing a backend URL that training workers interact with. Each environment follows a four-stage lifecycle:

\begin{itemize}
    \item \textbf{Task Reset.} At the beginning of each episode, the environment initializes the device state and loads a new task, ensuring a clean starting condition for every rollout.
    \item \textbf{Task Evaluation.} Virtual environments expose system-level root access, enabling reliable task completion verification through direct inspection of app state and database records. This system-level signal is further complemented by an MLLM-as-judge that assesses the final screen state against the task instruction, providing a robust and comprehensive outcome reward.
    \item \textbf{Spare Server Rotation.} Virtual sandbox environments are prone to becoming unhealthy during long training runs — a stalled or crashed container introduces training instability and can cause irrecoverable errors. To address this, ClawGUI-RL maintains a spare server queue. When a container is detected as unhealthy, the system automatically draws from the queue and rotates to a healthy replacement, allowing the affected task to resume without interrupting the training process.
    \item \textbf{Teardown.} Containers are periodically restarted to prevent state accumulation and maintain environment fidelity across long training runs.
\end{itemize}

\noindent\textbf{Real Device Training.} ClawGUI-RL supports training directly on physical Android devices or cloud phones through the same unified interface. Real-device training introduces two challenges that do not arise in virtual environments.

\begin{itemize}
    \item \textbf{Task Source.} Unlike virtual environments where tasks can be procedurally generated and automatically verified, real-device tasks must be manually authored to ensure they are both executable and verifiable on physical hardware. In ClawGUI-RL, we curate a set of human-authored tasks covering representative real-world scenarios.
    \item \textbf{Task Evaluation.} Physical devices do not expose system-level root access, making automated state verification infeasible. ClawGUI-RL therefore relies on MLLM-as-judge to assess task completion by evaluating the final screen state against the task instruction, providing a practical reward signal without requiring device-level privileges.
\end{itemize}

\subsubsection{Reward Design: Binary Reward and Dense Reward}

Reward design is critical for online RL training on long-horizon GUI tasks. ClawGUI-RL adopts a two-level reward formulation combining a binary outcome reward with a dense process reward.

\noindent\textbf{Binary Outcome Reward.} The primary reward signal is a binary score assigned at episode end: 1 for task success, 0 for failure. While straightforward, this signal suffers from a fundamental limitation in GUI environments — execution latency and multi-step interaction introduce significant delay between an action and its observable consequence, resulting in an extremely sparse reward signal that provides little guidance for intermediate steps.

\noindent\textbf{Dense Step-Level Reward via PRM.} To complement the sparse outcome reward, ClawGUI-RL integrates a Process Reward Model (PRM). After each action, the PRM receives the previous screenshot, the current screenshot, and the full history of actions taken so far, and judges whether the current action meaningfully contributes to task completion. This produces a per-step score that is combined with the outcome reward:
\begin{equation}
    R = R_{\text{outcome}} + R_{\text{step}}
\end{equation}
\noindent By providing dense feedback at every step, the PRM substantially alleviates the sparsity problem and enables the optimizer to distinguish productive actions from dead ends throughout the episode.

\subsubsection{RL Trainer}

ClawGUI-RL builds upon verl~\citep{Sheng_2025} and verl-agent~\citep{gigpo}, with
out-of-the-box support for a suite of RL algorithms including Reinforce++~\citep{reinforce}, PPO~\citep{schulman2017proximalpolicyoptimizationalgorithms}, GSPO~\citep{zheng2025groupsequencepolicyoptimization},
GRPO~\citep{shao2024deepseekmathpushinglimitsmathematical}, and
GiGPO~\citep{gigpo}. In our experiments, we integrate GRPO and GiGPO as the primary
advantage estimation algorithms and analyze their impact on GUI agent training.

\noindent GRPO~\citep{shao2024deepseekmath} estimates advantages by normalizing returns within a group of rollouts that share the
same task. While straightforward and effective for single-turn tasks, GRPO assigns a
uniform episode-level advantage to every step within a trajectory, which is too coarse for
long-horizon GUI interaction. Consider two rollouts on the same task: rollout $A$ completes
it in 4 steps, while rollout $B$ completes it in 8 steps. GRPO assigns both trajectories the
same reward, providing no signal to distinguish the efficiency of individual steps or to
credit productive actions over redundant ones.

\noindent GiGPO~\citep{gigpo} addresses this limitation through a two-level hierarchical advantage
estimation. At the episode level, GiGPO retains the macro relative advantage across complete
trajectories, preserving global trajectory quality signals. At the step level, GiGPO
introduces an anchor-state grouping mechanism: steps that encounter the same intermediate
environment state across different rollouts are retroactively clustered into sub-groups, and
micro relative advantages are estimated within each sub-group via discounted return
normalization. This hierarchical structure yields fine-grained per-step credit assignment
that captures both global trajectory quality and local step effectiveness, without requiring
a learned value network or additional rollouts, making it particularly well-suited to the
multi-step nature of GUI tasks.


\begin{figure}[t]
  \centering
  \includegraphics[width=0.98\linewidth]{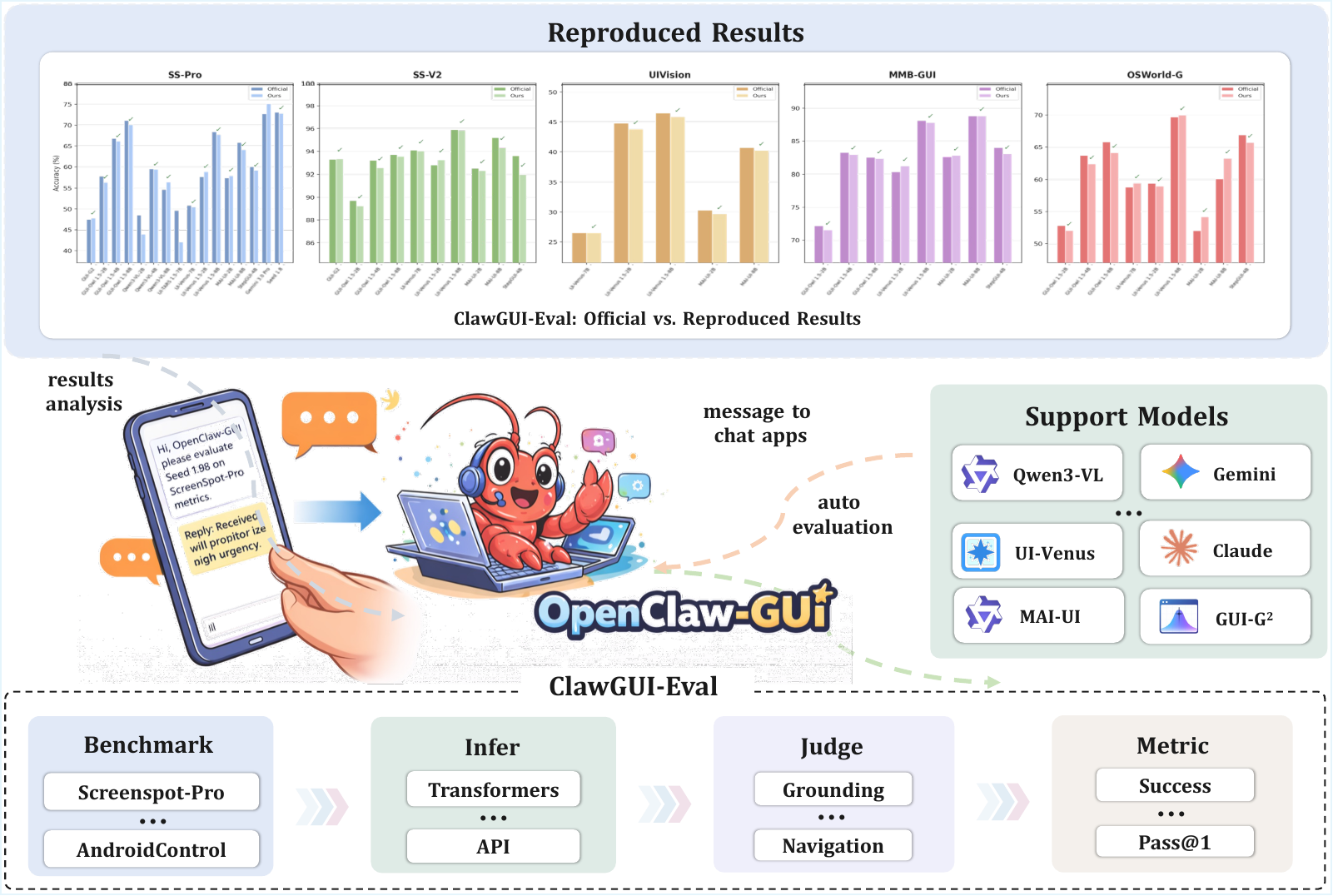}
\caption{Overview of ClawGUI-Eval, featuring a standardized Infer $\to$ Judge $\to$ Metric pipeline across 6 benchmarks and 11+ models. Reproduced results are compared against official baselines across five benchmarks, achieving a 95.8\% overall reproduction rate. The full pipeline can be triggered via a single natural language command through ClawGUI-Agent (OpenClaw-GUI).}
  \label{fig:clawgui-eval}
\end{figure}

\subsection{ClawGUI-Eval: Reproducible GUI Evaluation}

Evaluation is the compass of research progress, yet GUI evaluation is harder to reproduce than it appears. Prompt ordering, coordinate normalization conventions, image resolution, and sampling temperature interact in ways that shift reported accuracy by several points across implementations. As a result, numbers across papers are rarely comparable, and the community has no reliable baseline against which to measure true progress. As shown in Figure~\ref{fig:clawgui-eval}, ClawGUI-Eval addresses this by pinning all evaluation choices per model and adopting a strict three-stage pipeline, achieving a 95.8\% reproduction rate against official results across 6 benchmarks and 11+ models.

\subsubsection{Benchmark and Model Coverage}
ClawGUI-Eval covers 6 benchmarks spanning diverse GUI grounding and navigation scenarios: ScreenSpot-Pro~\citep{li2024screenspot-pro}, ScreenSpot-V2~\citep{wu2024osatlasfoundationactionmodel}, UI-Vision~\citep{uivision}, MMBench-GUI~\citep{mmbenchgui}, OSWorld-G~\citep{xie2025scalingcomputerusegroundinguser}, and AndroidControl~\citep{androidcontrol}. On the model side, it supports 11+ models including Qwen3-VL~\citep{qwen3vl}, Qwen2.5-VL~\citep{bai2025qwen25vltechnicalreport}, UI-TARS~\citep{qin2025uitarspioneeringautomatedgui}, MAI-UI~\citep{maiui}, GUI-G$^2$~\citep{tang2025gui}, UI-Venus~\citep{uivenus}, GUI-Owl~\citep{mobileagent3}, StepGUI~\citep{yan2025stepguitechnicalreport}, Gemini~\citep{Gemini}, and Seed 1.8~\citep{seed18}. All inference results are publicly released alongside the evaluation code, enabling the community to reproduce, extend, and build upon our results directly.

\subsubsection{Pipeline Architecture: Infer, Judge, and Metric}

ClawGUI-Eval decomposes evaluation into three decoupled stages, each with a clearly defined input and output.

\begin{itemize}
    \item \textbf{Infer.} Given a benchmark dataset and a target model, the inference stage generates raw predictions. ClawGUI-Eval supports two backends: local GPU inference via \texttt{transformers}, and remote API inference via any OpenAI-compatible endpoint. Multi-GPU parallel inference is handled automatically through Python \texttt{multiprocessing}, with each process pinned to a dedicated GPU. Shard-level checkpointing allows interrupted runs to resume from the last completed shard without recomputation.
    \item \textbf{Judge.} Raw model outputs are parsed and evaluated against ground truth. ClawGUI-Eval implements benchmark-specific judges: a point-in-box judge for standard GUI grounding benchmarks, a polygon and refusal-aware judge for OSWorld-G, and a multi-action judge for AndroidControl. Each judge produces a per-sample correctness label.
    \item \textbf{Metric.} Per-sample labels are aggregated into final accuracy scores with breakdowns by platform, UI element type, and task category, enabling fine-grained analysis beyond top-line numbers.
\end{itemize}

\noindent By decoupling these three stages, ClawGUI-Eval allows any single stage to be rerun independently — for instance, re-judging existing predictions with an updated parser without repeating expensive inference.

\begin{figure}[t]
  \centering
  \includegraphics[width=0.98\linewidth]{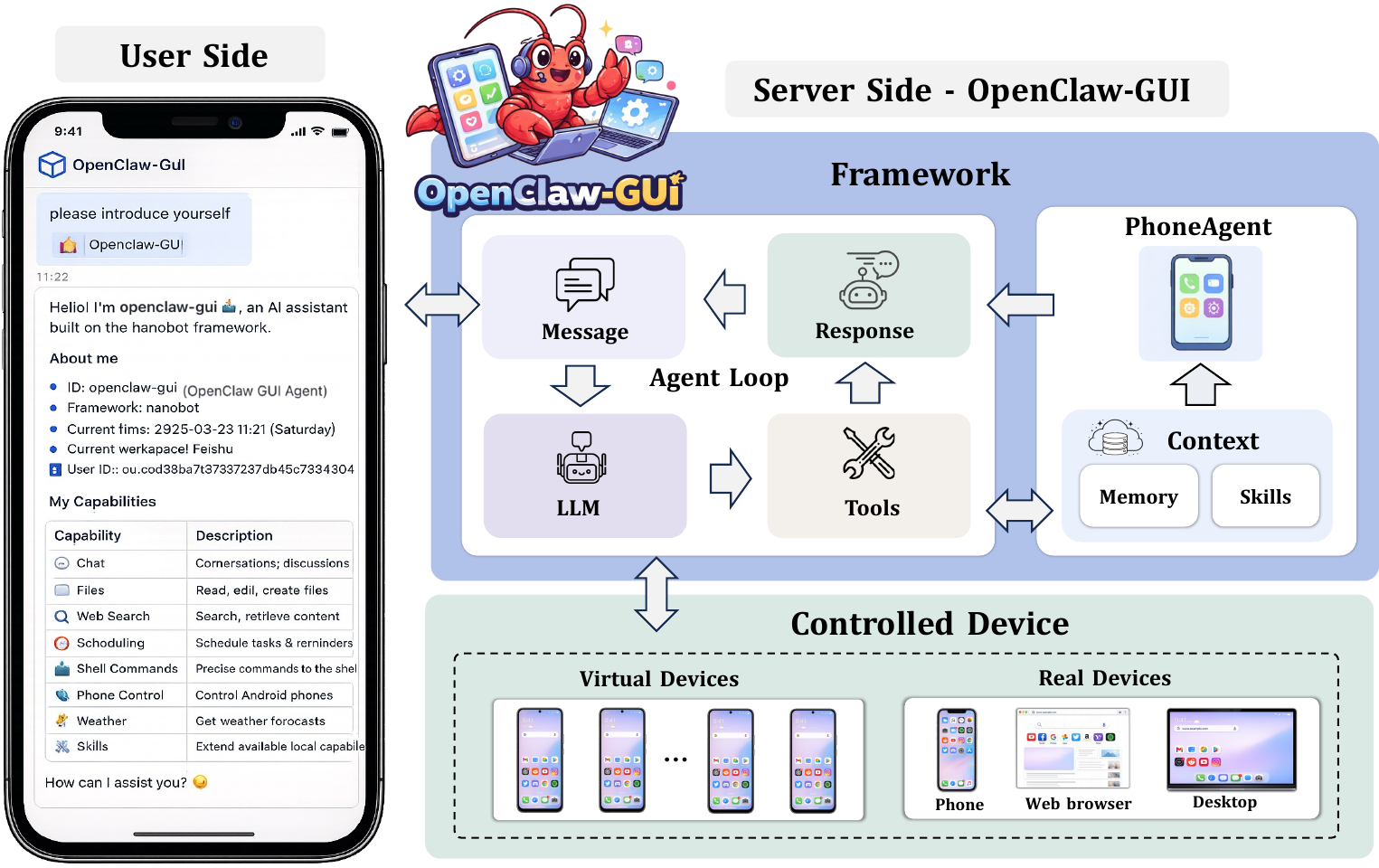}
\caption{Overview of ClawGUI-Agent, where users issue natural language instructions through 12+ chat platforms and the server side executes tasks via a message-driven agent loop with persistent memory and skills, controlling both virtual and 
real devices across phone, web browser, and desktop.}
  \label{fig:clawgui-agent}
\end{figure}

\subsection{ClawGUI-Agent: Personal GUI Assistant}

As GUI agents grow more capable, the final challenge is closing the loop to real users. A trained agent that cannot be deployed, personalized, or integrated into daily workflows delivers no practical value. As shown in Figure~\ref{fig:clawgui-agent}, ClawGUI-Agent is designed to bridge this gap, providing a production-ready system that brings GUI agents into the hands of real users across real devices.

\subsubsection{Hybrid Device Control: Operating via CLI and GUI}

Driven by the success of OpenClaw~\citep{openclaw}, a growing body of work has focused on CLI-based agent control. CLI interaction is precise and efficient: a single structured command can accomplish in one step what would otherwise require navigating multiple UI layers. However, CLI control carries fundamental limitations. Not all applications expose programmatic interfaces. CLI operations are opaque to users who cannot observe or intervene. It also bypasses the visual layer that makes agent behavior interpretable. GUI interaction addresses these limitations by operating directly on the screen, covering any application regardless of its underlying architecture, but introduces its own cost: tasks resolved by CLI in a single call may require multiple sequential GUI actions.

\noindent We argue that neither paradigm alone is sufficient. ClawGUI-Agent adopts a hybrid approach that leverages CLI efficiency where interfaces permit, and falls back to GUI control where they do not. This combination preserves the speed of CLI for well-supported operations while ensuring broad coverage through GUI for everything else.

\subsubsection{Personalized Memory}

ClawGUI-Agent incorporates a persistent personalized memory system. During task execution, the agent automatically extracts structured facts from interactions, including contact names and relationships, frequently used applications, and user habits and preferences, and stores them as vector embeddings in a persistent store. On subsequent tasks, the top-$k$ most semantically similar memories are retrieved and injected into the system context, allowing the agent to recognize recurring entities and adapt to individual user patterns over time. Duplicate memories are detected and merged rather than accumulated, keeping the memory store lean and relevant.

\subsubsection{Remote and Local Control}

ClawGUI-Agent supports two deployment modes. In \textbf{remote control} mode, the agent is accessed through 12+ chat platforms including Feishu, DingTalk, Telegram, Discord, Slack, and QQ, allowing users to issue tasks from a separate device to control the target phone remotely. In \textbf{local control} mode, users send instructions directly from a chat application running on the phone itself, upon which the agent takes over the local device in place, requiring no additional hardware or cloud relay.

\subsubsection{ClawGUI-Eval as a Deployable Skill}

ClawGUI-Agent exposes ClawGUI-Eval as a built-in tool skill, enabling users to trigger a complete benchmark evaluation pipeline through a single natural-language command. Upon receiving an instruction such as ``benchmark Qwen3-VL on ScreenSpot-Pro'', the agent automatically performs environment verification, launches multi-GPU parallel inference, runs the judge, computes metrics, and returns a structured result report with comparisons against official baselines, without writing a single script.


\begin{table}[t]
\centering
\begin{tabular}{lc}
\toprule
\textbf{Model} & \textbf{MobileWorld SR (GUI-Only)} \\
\midrule
\rowcolor{gray!15}
\multicolumn{2}{l}
{\textit{Agentic Framework}} \\
Claude-4.5-Sonnet + UI-Ins-7B & 47.8 \\
Gemini-3-Pro + UI-Ins-7B      & 55.6 \\
GPT-5 + UI-Ins-7B             & 54.0 \\
\midrule
\rowcolor{gray!15}
\multicolumn{2}{l}{\textit{End-to-End Model}} \\
GUI-Owl-7B         & 7.7  \\
GUI-Owl-32B        & 8.5  \\
UI-Venus-7B        & 8.5  \\
UI-Venus-72B       & 16.4 \\
Qwen3-VL-8B        & 9.4  \\
Qwen3-VL-32B       & 11.9 \\
Qwen3-VL-235B-A22B & 12.8 \\
Doubao-1.5-UI-TARS & 26.3 \\
MAI-UI-2B          & 11.1 \\
MAI-UI-8B          & 19.7 \\
\midrule
\rowcolor{gray!15}
\multicolumn{2}{l}{\textit{Ours}} \\
ClawGUI-2B         &  17.1 \\
\bottomrule
\end{tabular}
\caption{Comparison of models on GUI-Only (117 tasks) benchmark.}
\label{tab:gui_only}
\end{table}
\section{Experiments}
\subsection{Setting}
\textbf{Training.} We train ClawGUI-2B based on MAI-UI-2B~\citep{maiui} using 64 parallel virtual environments on 8$\times$A6000 (48GB) GPUs. We adopt the GiGPO algorithm with a rollout group size of 8, sampling temperature of 0.7, and learning rate of 1e-6 for 3 epochs with a training batch size of 8. For PRM-based step-level reward judgment, we employ Qwen3.5-72B as the judge model.

\noindent \textbf{Evaluation.} We evaluate ClawGUI-2B on MobileWorld, an online interactive benchmark designed to assess the end-to-end task completion capability of GUI agents. MobileWorld comprises three task categories: GUI-Only, MCP, and Call-User. We focus our evaluation on the GUI-Only split, which contains 117 tasks that require agents to complete real-world mobile interactions purely through visual GUI control, without any programmatic interface access. Task completion is measured by Success Rate, defined as whether the agent successfully accomplishes the task objective by the end of the episode. We set the maximum number of interaction steps to 50 for all evaluations.

\begin{table}[h]
\centering
\begin{tabular}{lcc}
\toprule
\textbf{Method} & \textbf{Reward Type} & \textbf{SR (\%)} \\
\midrule
GRPO  & Binary (episode-level)        & 14.5 \\
\rowcolor{gray!15}
GiGPO & Dense (episode- \& step-level) & \textbf{17.1} \\
\bottomrule
\end{tabular}
\caption{Ablation on reward design on MobileWorld GUI-Only (117 tasks).}
\label{tab:ablation_reward}
\end{table}
\subsection{Main Results.} Table ~\ref{tab:gui_only} reports Success Rate on the MobileWorld GUI-Only benchmark. We highlight three key observations.
(1) \textbf{Infrastructure drives policy quality}. ClawGUI-2B achieves 17.1\% SR, surpassing the same-scale MAI-UI-2B baseline by a relative margin of 6.0\%, which directly validates the effectiveness of our open-source RL infrastructure. Both models share the same base weights; the gain stems entirely from ClawGUI-RL's scalable environment management and reward design.
(2) \textbf{Small well-trained models outperform larger untrained ones.} ClawGUI-2B outperforms substantially larger end-to-end models, including Qwen3-VL-32B (11.9\%) and UI-Venus-72B (16.4\%), demonstrating that online RL training through real environment interaction contributes more to task completion capability than model scale alone.
(3) \textbf{Agentic frameworks remain a separate regime.} Methods that couple proprietary frontier models with dedicated grounding modules achieve higher absolute numbers (e.g., Gemini-3-Pro + UI-Ins-7B at 55.6\%), but rely on closed-source planners unavailable for end-to-end optimization. These systems are not directly comparable to compact trained agents and represent a complementary rather than competing paradigm.
Taken together, these results confirm that a well-engineered open-source training infrastructure can unlock strong GUI agent performance at modest model scale, closing a significant gap with much larger systems.

\subsection{Every Step Counts: Dense Reward Unlocks Better GUI Policies}

GRPO assigns a single advantage score to an entire episode, which is 
too coarse for long-horizon GUI tasks where intermediate steps vary 
significantly in quality. A misclick early in a trajectory receives 
the same credit as a decisive final action, providing little signal 
for the optimizer to distinguish productive steps from dead ends.

\noindent GiGPO addresses this through anchor-state grouping: steps that 
encounter the same intermediate state are clustered into sub-groups, and per-step advantages are estimated via discounted return  normalization within each sub-group. This yields fine-grained step-level credit assignment without requiring a learned value network, making it particularly well-suited to the multi-step nature 
of GUI interaction.

\noindent As shown in Table~\ref{tab:ablation_reward}, replacing GRPO with GiGPO yields a \textbf{2.6\% improvement} in SR (14.5\% $\rightarrow$ 17.1\%) on the MobileWorld GUI-Only tasks, a relative gain of 17.9\%. This consistent improvement demonstrates that dense step-level supervision provides a substantially richer learning signal than episode-level reward alone, and that accurate 
credit assignment is a critical factor in GUI agent training.


\begin{tcolorbox}[
    colback=zjublue!6!white,
    colframe=zjublue!70!white,
    colbacktitle=zjublue!85!white,
    coltitle=white,
    fonttitle=\bfseries,
    title=Takeaway
]
Dense step-level reward supervision via GiGPO yields a 17.9\% 
relative gain over episode-level GRPO, confirming that fine-grained 
credit assignment is a critical factor in GUI agent RL training.
\end{tcolorbox}

\subsection{Benchmarking the Benchmarks: Can We Trust Published GUI Numbers?}

Reproducibility is a prerequisite for meaningful progress, yet GUI evaluation is notoriously difficult to reproduce. Prompt ordering, coordinate normalization conventions, image resolution, and sampling temperature interact in ways that can shift reported accuracy by several points across implementations. As a result, numbers across papers are rarely directly comparable, and the community has lacked 
a reliable common baseline against which to measure true progress.

\noindent ClawGUI-Eval addresses this by pinning all evaluation choices per model and adopting a strict three-stage \textit{Infer $\rightarrow$ Judge $\rightarrow$ Metric} pipeline across 6 benchmarks and 11+ 
models. Table~\ref{tab:reproduction} reports our reproduced results against officially published numbers. A result is considered successfully reproduced ($\checkmark$) if the reproduced value meets or exceeds the official number, or the absolute difference is $\leq 2\%$.

\begin{table*}[t]
\centering
\resizebox{\textwidth}{!}{%
\begin{tabular}{lcccccccccc}
\toprule
\rowcolor{headerblue}
\textbf{Model} 
  & \textbf{SS-Pro Off.} 
  & \textbf{SS-Pro Ours} 
  & \textbf{SS-V2 Off.} 
  & \textbf{SS-V2 Ours}
  & \textbf{UIV Off.} 
  & \textbf{UIV Ours} 
  & \textbf{MMB Off.} 
  & \textbf{MMB Ours} 
  & \textbf{OSW-G Off.} 
  & \textbf{OSW-G Ours} \\
\midrule
\rowcolor{groupheader}
\multicolumn{11}{l}{\textit{Open-Source Models}} \\
\midrule
GUI-G$^2$        
  & 47.50 & \textcolor{successgreen}{\textbf{47.75}} 
  & 93.30 & \textcolor{successgreen}{\textbf{93.32}} 
  & \textcolor{gray}{-} & 25.99 
  & \textcolor{gray}{-} & 79.33 
  & \textcolor{gray}{-} & 58.63 \\
\rowcolor{rowgray}
GUI-Owl 1.5-2B   
  & 57.80 & \textcolor{successgreen}{\textbf{56.36}} 
  & 89.70 & \textcolor{successgreen}{\textbf{89.23}} 
  & \textcolor{gray}{-} & 23.71 
  & 72.17 & \textcolor{successgreen}{\textbf{71.54}} 
  & 52.80 & \textcolor{successgreen}{\textbf{52.04}} \\
GUI-Owl 1.5-4B   
  & 66.80 & \textcolor{successgreen}{\textbf{66.16}} 
  & 93.20 & \textcolor{successgreen}{\textbf{92.53}} 
  & \textcolor{gray}{-} & 29.97 
  & 83.24 & \textcolor{successgreen}{\textbf{82.94}} 
  & 63.70 & \textcolor{successgreen}{\textbf{62.34}} \\
\rowcolor{rowgray}
GUI-Owl 1.5-8B   
  & 71.10 & \textcolor{successgreen}{\textbf{70.08}} 
  & 93.70 & \textcolor{successgreen}{\textbf{93.55}} 
  & \textcolor{gray}{-} & 36.70 
  & 82.52 & \textcolor{successgreen}{\textbf{82.33}} 
  & 65.80 & \textcolor{successgreen}{\textbf{64.12}} \\
Qwen3-VL-2B      
  & 48.50 & \textcolor{failred}{43.90}                   
  & \textcolor{gray}{-} & 88.92 
  & \textcolor{gray}{-} & 15.06 
  & \textcolor{gray}{-} & 73.12 
  & \textcolor{gray}{-} & 54.12 \\
\rowcolor{rowgray}
Qwen3-VL-4B      
  & 59.50 & \textcolor{successgreen}{\textbf{59.39}} 
  & \textcolor{gray}{-} & 93.08 
  & \textcolor{gray}{-} & 27.78 
  & \textcolor{gray}{-} & 84.28 
  & \textcolor{gray}{-} & 68.43 \\
Qwen3-VL-8B      
  & 54.60 & \textcolor{successgreen}{\textbf{56.42}} 
  & \textcolor{gray}{-} & 94.26 
  & \textcolor{gray}{-} & 27.96 
  & \textcolor{gray}{-} & 84.25 
  & \textcolor{gray}{-} & 65.88 \\
\rowcolor{rowgray}
Qwen2.5-VL-3B    
  & \textcolor{gray}{-} & 15.62 
  & \textcolor{gray}{-} & 64.86 
  & \textcolor{gray}{-} & 6.73  
  & \textcolor{gray}{-} & 52.81 
  & \textcolor{gray}{-} & 26.08 \\
Qwen2.5-VL-7B    
  & \textcolor{gray}{-} & 27.45 
  & \textcolor{gray}{-} & 87.66 
  & \textcolor{gray}{-} & 14.40 
  & \textcolor{gray}{-} & 70.26 
  & \textcolor{gray}{-} & 35.49 \\
\rowcolor{rowgray}
UI-TARS 1.5-7B   
  & 49.60 & \textcolor{failred}{42.06}                   
  & \textcolor{gray}{-} & 89.54 
  & \textcolor{gray}{-} & 20.30 
  & \textcolor{gray}{-} & 73.23 
  & \textcolor{gray}{-} & 58.24 \\
UI-Venus-7B      
  & 50.80 & \textcolor{successgreen}{\textbf{50.47}} 
  & 94.10 & \textcolor{successgreen}{\textbf{94.03}} 
  & 26.50 & \textcolor{successgreen}{\textbf{26.52}} 
  & \textcolor{gray}{-} & 80.08 
  & 58.80 & \textcolor{successgreen}{\textbf{59.41}} \\
\rowcolor{rowgray}
UI-Venus 1.5-2B  
  & 57.70 & \textcolor{successgreen}{\textbf{58.82}} 
  & 92.80 & \textcolor{successgreen}{\textbf{93.24}} 
  & 44.80 & \textcolor{successgreen}{\textbf{43.82}} 
  & 80.30 & \textcolor{successgreen}{\textbf{81.19}} 
  & 59.40 & \textcolor{successgreen}{\textbf{58.97}} \\
UI-Venus 1.5-8B  
  & 68.40 & \textcolor{successgreen}{\textbf{67.68}} 
  & 95.90 & \textcolor{successgreen}{\textbf{95.83}} 
  & 46.50 & \textcolor{successgreen}{\textbf{45.88}} 
  & 88.10 & \textcolor{successgreen}{\textbf{87.79}} 
  & 69.70 & \textcolor{successgreen}{\textbf{69.98}} \\
\rowcolor{rowgray}
MAI-UI-2B        
  & 57.40 & \textcolor{successgreen}{\textbf{57.94}} 
  & 92.50 & \textcolor{successgreen}{\textbf{92.30}} 
  & 30.30 & \textcolor{successgreen}{\textbf{29.68}} 
  & 82.60 & \textcolor{successgreen}{\textbf{82.80}} 
  & 52.00 & \textcolor{successgreen}{\textbf{54.17}} \\
MAI-UI-8B        
  & 65.80 & \textcolor{successgreen}{\textbf{64.07}} 
  & 95.20 & \textcolor{successgreen}{\textbf{94.34}} 
  & 40.70 & \textcolor{successgreen}{\textbf{40.23}} 
  & 88.80 & \textcolor{successgreen}{\textbf{88.81}} 
  & 60.10 & \textcolor{successgreen}{\textbf{63.23}} \\
\rowcolor{rowgray}
StepGUI-4B       
  & 60.00 & \textcolor{successgreen}{\textbf{59.14}} 
  & 93.60 & \textcolor{successgreen}{\textbf{91.98}} 
  & \textcolor{gray}{-} & 29.90 
  & 84.00 & \textcolor{successgreen}{\textbf{83.03}} 
  & 66.90 & \textcolor{successgreen}{\textbf{65.69}} \\
\midrule
\rowcolor{groupheader}
\multicolumn{11}{l}{\textit{Closed-Source Models}} \\
\midrule
\rowcolor{closedrow}
Gemini 3.0 Pro (Zoom) 
  & 72.70 & \textcolor{successgreen}{\textbf{75.08}} 
  & \textcolor{gray}{-} & \textcolor{gray}{-} 
  & \textcolor{gray}{-} & \textcolor{gray}{-} 
  & \textcolor{gray}{-} & \textcolor{gray}{-} 
  & \textcolor{gray}{-} & \textcolor{gray}{-} \\
\rowcolor{closedrow}
Seed 1.8 (Zoom)       
  & 73.10 & \textcolor{successgreen}{\textbf{72.80}} 
  & \textcolor{gray}{-} & \textcolor{successgreen}{95.68} 
  & \textcolor{gray}{-} & \textcolor{gray}{-} 
  & \textcolor{gray}{-} & \textcolor{successgreen}{88.4} 
  & \textcolor{gray}{-} & \textcolor{gray}{-} \\
\rowcolor{closedrow}
Gemini 3.1 Pro (Zoom) 
  & \textcolor{gray}{-} & \textcolor{successgreen}{\textbf{85.01}} 
  & \textcolor{gray}{-} & \textcolor{gray}{-} 
  & \textcolor{gray}{-} & \textcolor{gray}{-} 
  & \textcolor{gray}{-} & \textcolor{gray}{-} 
  & \textcolor{gray}{-} & \textcolor{gray}{-} \\
\bottomrule
\end{tabular}}
\caption{Reproduction results across GUI grounding benchmarks. \textcolor{successgreen}{\textbf{Green bold}} indicates successful reproduction ($|\Delta| \leq 2\%$ or reproduced $\geq$ official); \textcolor{failred}{red} indicates a gap exceeding the threshold; \textcolor{gray}{-} indicates no official baseline. Benchmark abbreviations: \textbf{SS-Pro}~=~ScreenSpot-Pro, \textbf{SS-V2}~=~ScreenSpot-V2, \textbf{UIV}~=~UIVision, \textbf{MMB}~=~MMBench-GUI, \textbf{OSW-G}~=~OSWorld-G. Closed-source models are evaluated on ScreenSpot-Pro only via a two-stage Zoom paradigm.}
\label{tab:reproduction}
\vspace{-2mm}
\end{table*}

\noindent As shown in Table~\ref{tab:reproduction}, we highlight three observations. First, ClawGUI-Eval achieves an \textbf{overall reproduction rate of 95.8\%} (46/48 cells with official baselines), with open-source models reaching 95.7\% and frontier models reaching 100\% on ScreenSpot-Pro. Second, the two failure cases (Qwen3-VL-2B and UI-TARS 1.5-7B on SS-Pro) both involve models whose official evaluation configurations have not been publicly disclosed, suggesting that undisclosed prompt or resolution choices are the primary driver of irreproducibility in the field. Third, for closed-source frontier models where standard inference is infeasible~\citep{why_your_ai_agent_keeps_misclicking}, we adopt a \textbf{Zoom paradigm}, a two-stage crop-then-ground strategy that applies 25\% crop tiles for Gemini and 50\% crop tiles for Seed, which successfully recovers official performance without any access to model internals.

\begin{tcolorbox}[
    colback=zjublue!6!white,
    colframe=zjublue!70!white,
    colbacktitle=zjublue!85!white,
    coltitle=white,
    fonttitle=\bfseries,
    title=Takeaway
]
A standardized pipeline with pinned evaluation choices achieves 
95.8\% reproduction rate across 6 benchmarks and 11+ models, 
demonstrating that GUI evaluation discrepancies are an 
infrastructure problem, not a fundamental limitation.
\end{tcolorbox}


\section{Discussion}

\paragraph{Toward a Unified GUI-CLI Agentic Harness.}

The dominant lesson from the past year of agent engineering, from Claude Code~\citep{claudecode,anthropic_harness_design_2026} and Hermes Agent~\citep{hermes_agent} to MiniMax M2.7's self-evolving loop~\citep{minimax_m27_news_2026}, is that frontier capability comes as much from the surrounding harness as from the model itself. Permission pipelines, tool dispatch, context compaction, and multi-turn recovery determine whether a capable model becomes a reliable agent or spirals after several steps~\citep{anthropic_harness_design_2026}. Yet CLI-based~\citep{openclaw,hermes_agent} and GUI-based~\citep{uitars2,uivenus15,maiui} agents have grown as two parallel ecosystems with almost no shared infrastructure, despite overlapping user goals. Recent hybrid systems such as Hermes Agent's unified terminal-to-Android gateway~\citep{hermes_agent} and MiniMax's shell-browser-MCP toolchains~\citep{minimax_m27_news_2026} hint at convergence. We view ClawGUI as an early step toward a shared harness standard that treats CLI, GUI, and API calls as interchangeable actions and learns the routing policy itself from interaction data.

\paragraph{Scaling Online RL Beyond Emulators.}
Current GUI agent RL training is confined almost entirely to emulator sandboxes~\citep{mobileguirl,uitars2,uivenus15,mobileworld}, which drift from real app behavior and cannot cover the authenticated long tail of commercial applications. Two complementary directions are emerging. First, mock applications reconstructed by modern code-generation models~\citep{minimax_m27_news_2026,claudecode} offer an authentication-free distribution that mirrors real interaction flows without real user credentials. Second, on-device RL with privacy-preserving trajectory collection can tap the enormous pool of real user interaction without centralizing data. Both directions assume an infrastructure that handles environment instability at scale, which is precisely the role ClawGUI-RL is designed to fill. Scaling online RL is now as much a systems problem as an algorithmic one.

\paragraph{Toward On-Device, Always-Present System Agents.}
As on-device inference becomes practical, the final shape of a GUI agent looks less like a remote service invoked on demand and more like a persistent system-level intelligence running locally. Recent efforts such as Hermes Agent's Android device control~\citep{hermes_agent} and Google's Gemma 4, an effective 2B mobile-first model~\citep{google_gemma4_blog_2026}, together suggest that capable agentic models and ubiquitous device-level deployment are now converging. Such an agent would perceive full device state, retain persistent personalized memory, and execute multi-app workflows autonomously in the background. ClawGUI-Agent's hybrid CLI-GUI control and personalized memory system are early instances of this pattern, but the full vision requires tighter operating-system integration, on-device policy training, and rigorous local-first privacy guarantees that the community has yet to establish.

\paragraph{World Models for GUI Environments.}
Today's GUI agents act reactively: observe a screenshot, predict an action, wait for environment feedback. What they lack is an internal model of how the screen will evolve in response to a candidate action, which is precisely what allows humans to plan several steps ahead before committing. Recent progress on general-purpose world models~\citep{zheng2026code2worldguiworldmodel,luo2025vimogenerativevisualgui,deepmind_genie3_blog_2025} suggests that learning UI dynamics as a predictive model is now tractable, with training signals drawn from the same screen-action trajectories already collected for imitation and RL. A GUI-specific world model would enable model-based planning, counterfactual rollouts, and early dead-end detection, turning multi-step interaction from blind trial-and-error into deliberate search. We view ClawGUI-RL's dense step-level trajectory logging as a natural substrate for training such models at scale.

\section{Conclusion}

We presented ClawGUI, a unified open-source framework that integrates online RL training, standardized evaluation, and real-device deployment into a single coherent pipeline for GUI agent development. ClawGUI-RL provides the first open-source infrastructure with validated support for both large-scale parallel virtual environments and real physical device training, integrating GiGPO with a Process Reward Model for dense step-level reward supervision. ClawGUI-Eval establishes a reproducible evaluation standard across 6 benchmarks and 11+ models, achieving a 95.8\% reproduction rate against official 
baselines. ClawGUI-Agent closes the loop from research to deployment, enabling natural-language-driven automation across Android, HarmonyOS, and iOS through 12+ chat platforms with a personalized memory system. Trained end-to-end within this pipeline, ClawGUI-2B achieves 17.1\% MobileWorld SR, outperforming the same-scale baseline by a relative margin of 54\% and surpassing models of substantially larger scale. We hope ClawGUI serves as a foundation for the community to build, evaluate, and deploy the next generation of GUI agents.










\bibliography{references}


\end{document}